\documentclass[conference]{IEEEtran}
\IEEEoverridecommandlockouts

\usepackage{amsmath,amssymb,amsfonts}
\usepackage{algorithm}
\usepackage{algpseudocode}
\usepackage{graphicx}
\usepackage{textcomp}
\usepackage{xcolor}
\usepackage{booktabs}
\usepackage{multirow}
\usepackage{ulem}
\usepackage{subcaption}
\usepackage[table]{xcolor}

\def\BibTeX{{\rm B\kern-.05em{\sc i\kern-.025em b}\kern-.08em
    T\kern-.1667em\lower.7ex\hbox{E}\kern-.125emX}}

\begin{document}

\title{Parameter-Efficient Fine-Tuning for HAR: Integrating LoRA and QLoRA into Transformer Models}

\author{\IEEEauthorblockN{ Irina SEREGINA}
\IEEEauthorblockA{Univ. Grenoble Alpes,\\ Grenoble, France \\
Irina.Seregina@etu.univ-grenoble-alpes.fr}
\and

\IEEEauthorblockN{Philippe LALANDA}
\IEEEauthorblockA{Univ. Grenoble Alpes,\\ Grenoble, France \\
philippe.lalanda@imag.fr}
\and
\IEEEauthorblockN{German VEGA}
\IEEEauthorblockA{Univ. Grenoble Alpes,\\ Grenoble, France \\
german.vega@imag.fr}

}
\maketitle

\begin{abstract}

Human Activity Recognition (HAR) is a foundational task in ubiquitous computing with applications in health monitoring, smart environments, and human–computer interaction. While recent advances in self-supervised learning and transformer-based architectures have significantly improved HAR performance, adapting large pretrained models to new domains remains a practical challenge due to limited computational resources on target devices. This papers investigates parameter-efficient fine-tuning techniques, specifically Low-Rank Adaptation (LoRA) and Quantized LoRA, as scalable alternatives to full model fine-tuning for HAR. We propose an adaptation framework built upon a Masked Autoencoder backbone and evaluate its performance under a Leave-One-Dataset-Out validation protocol across five open HAR datasets. Our experiments demonstrate that both LoRA and QLoRA can match the recognition performance of full fine-tuning while significantly reducing the number of trainable parameters, memory usage, and training time. Further analyses reveal that LoRA maintains robust performance even under limited supervision and that the adapter rank provides a controllable trade-off between accuracy and efficiency. QLoRA extends these benefits by reducing the memory footprint of frozen weights through quantization, with minimal impact on classification quality.

\end{abstract}

\maketitle

\section{Introduction} \label{sec:intro}

Deep learning has become a cornerstone of modern artificial intelligence, particularly in pervasive computing applications \cite{pervasiveTrend}. Unlike traditional machine learning approaches that rely on handcrafted features, deep models automatically extract representations from raw sensor time series, making them especially effective for high-dimensional and heterogeneous data. A major breakthrough was the introduction of Transformers—self-attention architectures that capture long-range dependencies without recurrence \cite{vaswani2017attention}. Originally designed for natural language processing, they have since been adapted to sensor-based time-series tasks \cite{ek2023transformer}. Their scalability and representational power make them attractive for pervasive computing; however, they typically require large training datasets to converge and involve millions of parameters, which limits their deployment on resource-constrained devices. Moreover, in pervasive computing, labeled data are often scarce due to costly annotation and high context variability, further exacerbating the data-hungry nature of Transformers.


To mitigate label scarcity, practitioners increasingly adopt a two-stage pipeline: pre-training on large, heterogeneous corpora followed by task-specific fine-tuning. Pre-training yields transferable, general-purpose features, while fine-tuning adapts them to downstream tasks with limited supervision. This approach has proved effective in many pervasive applications—for example, self-supervised pre-training on large wearable collections before Human Activity Recognition (HAR) adaptation \cite{ek2024comparing}, or pre-training on multi-sensor sleep datasets prior to specialization on smaller clinical cohorts \cite{kawabata2025switching}. Beyond representation transfer, pre-training effectively enlarges the usable training signal by leveraging related—though not identical—domains and contexts. This label efficiency is especially valuable for Transformers, whose strong capacity entails substantial data requirements.


The purpose of fine-tuning is to adapt a (pretrained) model to a target domain by updating its parameters. Full fine-tuning, where all parameters are adjusted, often yields strong performance but comes at high computational and memory costs, which limits its applicability to resource-constrained devices. To address this, parameter-efficient fine-tuning (PEFT) methods have been proposed. Techniques such as Low-Rank Adaptation (LoRA) or adapters update only a small fraction of additional parameters, while keeping most pretrained weights frozen. This approach reduces training and storage costs substantially while preserving accuracy and reducing the amount of labeled data required for adaptation, making it attractive for deployment in real-world settings.



In this paper, we study human activity recognition and incorporate Low-Rank Adaptation (LoRA) into a Transformer backbone pretrained via Masked Autoencoding (MAE). We show that this integration is both feasible and effective, offering a practical path to scalable, low-cost—and data-efficient—personalization in HAR and, more broadly, pervasive computing. The choice of HAR as our application domain is motivated by the growing need for systems that robustly adapt to non-stationary, real-world conditions. As deployments scale across users, devices, and contexts, HAR models must deliver accurate, personalized predictions on resource-constrained mobile and wearable platforms---often with limited labeled data and under strict privacy constraints.
Conventional supervised models, even when pretrained, often fail to generalize without costly full-model adaptation. We further extend the study to QLoRA, which introduces quantization to reduce memory footprint and improve inference speed, enabling practical deployment on edge and IoT devices.  

More specifically, we investigate whether integrating Transformers and (Q)LoRA fine-tuning enhances the adaptability of HAR models under domain shifts and low-resource conditions.  We benchmark the proposed approach on multiple publicly available HAR datasets and evaluate its performance using a Leave-One-Dataset-Out validation protocol across five heterogeneous HAR benchmarks.



The key contributions of this work are then as follows: 
\begin{itemize}  
    \item We introduce the integration of LoRA into transformer-based HAR models pretrained with MAE, enabling parameter-efficient fine-tuning for domain adaptation.  
    \item We extend this approach with QLoRA, reducing the memory footprint of frozen weights through quantization and enabling faster, more resource-efficient deployment.  
    \item We conduct extensive experiments under a Leave-One-Dataset-Out validation protocol across five heterogeneous HAR datasets, demonstrating that LoRA maintains robust performance while providing a tunable trade-off between accuracy and efficiency via adapter rank. QLoRA further enhances deployability by lowering resource requirements with minimal impact on accuracy.  
\end{itemize}

The paper is structured as follows: Section 2 reviews related work on HAR, and fine-tuning strategies with a focus on parameter-efficient adaptation, including LoRA and QLORA. Section 3 describes the proposed methodology, including the model architecture, the integration of LoRA and QLoRA, and dataset preprocessing. Section 4 describes the implementation and section V presents experimental results and comparative analyses between full fine-tuning, LoRA, and QLoRA. Finally, Section 5 concludes the paper with a summary of findings and directions for future work.

\section{Background and related work} \label{sec:preliminsries}

\subsection{Human Activity Recognition}

Human Activity Recognition (HAR) is the automatic identification of physical activities from data collected by wearable or ambient sensors such as accelerometers and gyroscopes cite \cite{jobanputra2019human, Stisen2015}. It supports applications in healthcare, personalized fitness, rehabilitation, workplace safety, and smart homes. Over the years, HAR has progressed from shallow classifiers with handcrafted features to deep learning models such as CNNs, RNNs, and transformers, which can learn spatiotemporal representations directly from raw multivariate signals~\cite{Khan2021DeepHARsurvey}. Yet HAR suffers from pronounced domain shift: models trained on one dataset often degrade on unseen users, devices, placements, sampling rates, or behaviors~\cite{dataset2023har}. In HAR, domain shifts commonly arise due to:
\begin{itemize}
    \item Covariate shift: the input distribution $P(x)$ changes while $P(y\!\mid\!x)$ remains the same (e.g., older adults walking more slowly than younger adults);
    \item Label shift: the label distribution $P(y)$ changes while $P(x\!\mid\!y)$ remains the same (e.g., datasets dominated by low- vs.\ high-intensity activities);
    \item Concept shift: the relationship $P(y\!\mid\!x)$ changes (e.g., identical motion patterns labeled differently across contexts).
\end{itemize}
Hence, even with strong pretrained representations, fine-tuning is fundamental to adapt models to the target domain (user, device, placement) and to recover robustness under heterogeneity and noise.

\subsection{Fine Tuning: basics}

Full fine-tuning is the process of updating \emph{all} parameters of a pretrained neural network when adapting it to a new task or domain (as usually defined in SE \cite{4375240,4724590}). It remains the standard and most flexible method for adapting large models, especially when there is a significant mismatch between the source and target data distributions. A typical full-adaptation pipeline is: (1) initialize a pretrained backbone (general-purpose representations); (2) add a task-specific head (e.g., a softmax classifier for HAR); (3) unfreeze all layers (as opposed to feature extraction); and (4) train end-to-end on the target data (e.g., with cross-entropy), so gradients update the entire network and both low- and high-level features adapt.

Full fine-tuning is particularly effective when the target task differs significantly from pretraining, often yielding higher accuracy in transfer scenarios with small datasets or strong domain shifts~\cite{fullfinetune2020,parthasarathy2024ultimate}, because updating all layers can realign both low- and high-level features to the new input–label distribution. It maximizes domain-specific learning capacity but at the expense of higher compute, sensitivity to initialization, and risk of forgetting pretrained knowledge. However, full fine-tuning is rarely suited for edge deployment: updating all weights is memory- and compute-intensive, and storing a separate model for each user or context makes large-scale personalization impractical.  

\subsection{Fine Tuning: Advanced techniques}

In addition to full fine-tuning, a variety of alternative strategies have been developed to adapt pretrained models to new tasks or domains. These methods are particularly relevant in scenarios where computational resources are limited or where parameter efficiency is essential. In particular, let us mention the following approaches:

\begin{itemize}
    \item Feature Extraction (Frozen Backbone): 
In this approach, the pretrained model is used as a fixed feature extractor. All weights in the backbone are frozen, and only the newly added task-specific head is trained on a new data source. This strategy is computationally efficient and requires minimal memory, making it well-suited for low-resource settings or when the target dataset is small. However, its adaptability is limited because the frozen backbone cannot adjust to the characteristics of the new data distribution.

    \item Partial fine-tuning refers to the selective unfreezing and training of only a subset of model parameters, typically the top layers of the encoder or normalization layers. This technique seeks a balance between performance and resource usage. It allows the model to adapt its high-level representations to the new domain while preserving the general features learned during pretraining.

    \item  Adapter layers introduce small, trainable modules into the frozen pretrained model. These modules are typically lightweight feed-forward networks inserted within the transformer architecture, often between attention and feed-forward blocks \cite{adapters2019}. During the training phase, only the parameters of these adapter modules are updated, while the rest of the model remains unchanged.

\end{itemize}

Other fine-tuning approaches include:
\begin{itemize}
    \item BitFit — updates only the bias terms of a neural network, achieving competitive performance in some NLP tasks.
    \item Prompt and Prefix Tuning — adapts transformer models by learning trainable prompt or prefix vectors.
    \item Hypernetwork-Based Tuning — uses a smaller auxiliary neural network to generate task-specific weights for the main model.
\end{itemize}

These techniques provide a rich set of tools for model adaptation. However, each of them presents limitations when applied to the HAR domain and IoT-focused deployment scenarios. BitFit significantly reduces the number of trainable parameters but offers limited expressive capacity, especially for non-language tasks like time-series sensor classification, where bias-only updates are unlikely to capture complex domain shifts. Prompt and Prefix Tuning were primarily designed for autoregressive language models and rely on manipulating input embeddings or attention keys; this mechanism does not directly translate to time-series transformers, where input semantics and temporal structure differ significantly from NLP settings. Hypernetwork-based tuning, while flexible and effective in some multitask scenarios, introduces additional computational overhead by requiring a separate network to generate weights, which contradicts the memory and runtime constraints typical of embedded HAR systems.


\subsection{LoRA}

Low-Rank Adaptation (LoRA) \cite{OriginLORA2021} is a parameter-efficient fine-tuning technique in which additional low-rank trainable matrices are inserted into existing layers, rather than updating all original parameters.
Formally, for a linear map with weight \(W\in\mathbb{R}^{d_{\text{out}}\times d_{\text{in}}}\), LoRA parameterizes a low-rank update \(\Delta W = BA\) with rank \(r \ll \min(d_{\text{out}}, d_{\text{in}})\), where \(B\in\mathbb{R}^{d_{\text{out}}\times r}\) and \(A\in\mathbb{R}^{r\times d_{\text{in}}}\).
The forward pass becomes
\[
y \;=\; W x \;+\; \frac{\alpha}{r}\, BAx,
\]
keeping \(W\) frozen and training only \(A,B\) (typically with \(A\) initialized to zero so the initial function is unchanged). At inference time, the update can be merged via
\[
W \;\leftarrow\; W \;+\; \frac{\alpha}{r}\, BA.
\]
This approach substantially reduces the number of trainable weights while maintaining performance close to that of full fine-tuning.
Recent surveys discuss the various design choices underlying LoRA, including the selection of insertion points for the adapters, the choice of rank, and the resulting trade-offs between accuracy and efficiency \cite{LoRASurvey2024,PEFTSurvey2024}.
Comparative studies further indicate that LoRA and full fine-tuning converge to different parameter configurations, which may explain differences in their ability to generalize to unseen data \cite{LoRAIllusion2024}.


Although LoRA was originally introduced for large language models, it has since gained popularity in computer vision \cite{CV2024VLoRA,CV2025ACLoRA}, speech processing \cite{ASR2024LoRAWhisper,SR24SparseWhisper}, and time series forecasting \cite{Gupta2024TimeSeriesLoRA,TimeSeries2024STLoRA}. LoRA is modular, meaning it can be added or removed from a pretrained model without altering the original weights, and it has a low memory footprint, requiring very little additional storage for the trainable parameters. These properties make it especially appealing for HAR. 


\subsection{QLoRA}

QLoRA extends LoRA by combining it with 4-bit quantization for base model weights (using the NF4 quantization format), keeping LoRA matrices in higher precision. This reduces memory usage and allows fine-tuning of large models on a single consumer-grade GPU without significant accuracy loss. Key innovations include NF4 quantization, double quantization to compress quantization constants, and paged optimizers to reduce memory peaks during backpropagation. It has been demonstrated that QLoRA  can train very large models to state-of-the-art quality while using much less memory. A well-known example is the Guanaco model family, which showed top results on the Vicuna-bench benchmark and, in some cases, matched or even beat models fine-tuned in full 16-bit precision, but with a fraction of the hardware requirements \cite{QLoRA2023}.

IR-QLoRA \cite{IRQLoRA2024} replaces certain operations with integer-only arithmetic to improve stability when working with quantized weights. Another method, LoftQ \cite{LoftQ2023}, integrates the quantization process with LoRA fine-tuning so that the two steps work together, which can improve accuracy and sometimes even outperform standard QLoRA. These developments aimed at making quantized fine-tuning more stable and more accurate, especially for tasks where every bit of memory and compute power matters.

While direct applications of QLoRA to HAR are not yet common, parameter-efficient fine-tuning with quantization has been actively explored in time-series modeling \cite{TimeSeriesLoRA2024}, making QLoRA a natural candidate for future HAR systems. For resource-constrained HAR scenarios, QLoRA provides an interesting trade-off between performance and efficiency, enabling scalable personalization in IoT deployments.

\begin{table*}[h]
\footnotesize
\caption{Summary of datasets characteristics}
\label{tab:datasets}
\centering
\scalebox{0.85}{
\begin{tabular}{@{}c|c|c|c|c|c|c@{}}
\toprule
\textbf{Dataset} &
  \textbf{\begin{tabular}[c]{@{}c@{}}\# of \\ samples\end{tabular}} &
  \textbf{\begin{tabular}[c]{@{}c@{}}\# of \\ users\end{tabular}} &
  \textbf{Adopted Devices} &
  \textbf{Sampling rate} &
  \textbf{Device position} &
  \textbf{Activities} \\ \midrule
HHAR &
 85,567 &
  9 &
  \begin{tabular}[c]{@{}c@{}}Smartphones: Samsung Galaxy S3 mini,\\  Samsung Galaxy S3, LG Nexus 4,\\  Samsung Galaxy S+\\ \\ Smartwatches: LG watches, \\ Samsung Galaxy Gears\end{tabular} &
  \begin{tabular}[c]{@{}c@{}}from 50 Hz \\ to 200 Hz\end{tabular} &
  \begin{tabular}[c]{@{}c@{}}Smartphones: Waist \\ \\ Smartwatches: Wrist\end{tabular}
  &
  \begin{tabular}[c]{@{}c@{}}Biking, Sitting, Standing,\\ Walking, Upstairs, Downstairs\end{tabular} \\ \midrule

MotionSense &
  17,231  &
  24 &
  Apple iPhone 6s &
  50 Hz &
  \begin{tabular}[c]{@{}c@{}}Waist\end{tabular} &
  \begin{tabular}[c]{@{}c@{}}Downstairs, Upstairs, Sitting, \\ Standing, Walking, Running\end{tabular} \\ \midrule
RealWorld &
  356,427 &
  15 &
  \begin{tabular}[c]{@{}c@{}}Samsung Galaxy S4 \\ \\ LG G Watch R\end{tabular} &
  50 Hz &
  \begin{tabular}[c]{@{}c@{}}Smartphones: Head, Chest, Upper arm, \\ Waist, Thigh, Shin  \\ \\ Smartwatches: Forearm\end{tabular} &
  \begin{tabular}[c]{@{}c@{}}Downstairs, Upstairs, Lying, \\ Sitting, Standing, Jumping, \\ Walking, Running\end{tabular} \\ \midrule
UCI &
  10,299 &
  30 &
  Samsung Galaxy S II &
  50 Hz &
  Waist &
  \begin{tabular}[c]{@{}c@{}}Walking, Upstairs, Downstairs, \\ Sitting, Standing, Lying\end{tabular} \\ \midrule
PAMAP2 &
  15,177 &
  8 &
  Colibri wireless IMU sensors &
  100 Hz &
  Waist, Chest, Wrist &
  \begin{tabular}[c]{@{}c@{}}Rope Jumping, Lying, Sitting,\\  Standing, Walking, Running, \\ Cycling, Nordic walking, \\ Upstairs, Downstairs,\\ Vacuum cleaning, Ironing\end{tabular} \\ \bottomrule
\end{tabular}
}

\end{table*}

\section{Approach} \label{sec:method}

\subsection{Overview}



The objective of this work is to evaluate three fine-tuning strategies for sensor-based HAR—full fine-tuning, LoRA, and QLoRA—and to quantify their trade-offs in recognition accuracy, compute, and memory when adapting a high-capacity model to new domains. We ground the study in a state-of-the-art Transformer backbone pretrained with a Masked Autoencoder (MAE), a strong choice for multivariate time series \cite{ek2023transformer}. Our deployment target is not ultra-constrained microcontrollers, but edge-class devices—smartphones, smartwatches, and embedded gateways—that can execute Transformer inference yet offer limited headroom for on-device training (RAM/VRAM, bandwidth, and battery). 
Given the backbone’s size, efficient adaptation is therefore both challenging and necessary to meet these edge constraints.

Also, all methods are evaluated under a common protocol, ensuring that the comparison is both rigorous and representative of realistic deployment conditions. To our knowledge, this is the first systematic study in HAR that simultaneously implements and benchmarks full fine-tuning, LoRA, and QLoRA on the same backbone, thereby offering novel insights into their relative merits.  



\subsection{Pre-training Strategy}

Training a model on a single dataset rarely captures the diversity of sensing conditions across devices, placements, and user populations, often resulting in poor generalization. To address this, we adopt cross-dataset pretraining on five widely used HAR datasets (see Table~I):

\subsubsection{Heterogeneity Human Activity Recognition (HHAR) \cite{10.1145/2809695.2809718}}
9 participants wearing 8 smartphones (waist pouch) and 4 smartwatches performed 6 activities. Accelerometer and gyroscope were sampled at 50–200\,Hz across 12 heterogeneous devices.

\subsubsection{MotionSense \cite{Malekzadeh:2018:PSD:3195258.3195260}}
24 subjects carried an iPhone~6s in the front pocket; accelerometer, gyroscope, and attitude were recorded at 50\,Hz over 6 activities (walking, jogging, stairs, sitting, standing). Controlled protocol with phone-only sensing.

\subsubsection{RealWorld \cite{realword}}
15 subjects (18\,h total) with a Galaxy S4 and LG G Watch~R at 7 body locations collected accelerometer/gyroscope at 50\,Hz over 8 activities in unconstrained outdoor settings. Notable class imbalance (e.g., standing vs.\ jumping) and cross-position variability.

\subsubsection{UCI Human Activity Recognition \cite{Anguita2013APD}}
30 subjects wore a Galaxy S~II on the waist (50\,Hz) to perform 6 activities in a controlled lab. A canonical benchmark with low device variability and well-defined conditions.

\subsubsection{PAMAP2 \cite{reiss2012introducing}}
9 subjects performed 12 daily/exercise activities wearing three IMUs (ankle, chest, wrist) with accelerometer/gyroscope/magnetometer at 100\,Hz. Controlled setup with multi-sensor, multi-location recordings.

We downsampled all datasets to 50 Hz, consistent with evidence that 20–50 Hz is optimal for smartphone HAR and that accelerometer/gyroscope suffice; higher rates add cost with marginal gains~\cite{overviewHar}. Each dataset was sensor-wise z-normalized independently to prevent small corpora being dominated by larger ones (e.g., UCI vs. HHAR). To minimize position-induced domain shift and focus on data scarcity, we retained only waist-mounted recordings present across datasets. Signals were segmented into 128-sample (2.56 s) windows with 50\% overlap over the six accelerometer/gyroscope channels~\cite{ignatov2018real}. Datasets were then combined by taking the union of activity labels.

\subsection{Evaluation Strategy}

In order to evaluate fine tuning, we use a  strategy called "Leave-One-Dataset-Out" (LODO) \cite{dataset2023har}. Precisely, At each fold, one dataset is considered as \textit{left-out dataset}, while the \textit{remaining} ones are used to create a pre-trained model. 

Once pretrained, the encoder is fine-tuned on the dataset that was excluded during pretraining. This protocol is rotated across all combinations, ensuring that each dataset serves as the target domain exactly once. The LODO design provides a principled and rigorous way to assess cross-dataset transfer in HAR. Unlike random within-dataset splits, which primarily test interpolation, LODO explicitly measures extrapolation to unseen sources—a scenario that closely mirrors real-world IoT deployments where models must adapt to new hardware, environments, or user populations. This strategy therefore offers a comprehensive assessment of how different fine-tuning strategies perform under realistic domain shift conditions.

Full fine-tuning serves as the baseline: it updates all parameters of the pretrained MAE and thus provides an upper bound on recognition accuracy. Against this baseline, we compare LoRA and QLoRA. These methods are evaluated under the LODO protocol described above, ensuring that the comparison is both rigorous and representative of realistic deployment conditions. 


\section{Implementation}

\subsection{MAE and Transformer-based architecture}

In our framework, the backbone model is a Transformer encoder–decoder architecture, while the training pipeline follows the Masked Autoencoder (MAE) paradigm. 
It is important to emphasize this distinction: the Transformer specifies the architectural building blocks (attention layers, feed-forward networks, residual connections), whereas MAE defines the self-supervised learning strategy (masking input patches and reconstructing them). 
Together, they provide a powerful combination for learning robust spatiotemporal representations from multivariate sensor data.

\paragraph{Input representation and patching}
Let $X \in \mathbb{R}^{L \times C}$ denote a sensor window of length $L$ with $C$ channels. 
The MAE pipeline first partitions the signal into non-overlapping patches of length $P$, producing $\tfrac{L}{P}$ patches per channel. 
Each patch is projected into a $d$-dimensional embedding and augmented with positional encodings to retain temporal order:
\[
z_i = \mathrm{Linear}(X_{i:i+P}) + p_i, \qquad z_i \in \mathbb{R}^d .
\]

During pretraining, a fraction $m = 75\%$ of patch tokens is masked at random. 
The encoder processes only the visible tokens, while the decoder receives the concatenation of encoded visible tokens and learned mask tokens at masked positions. 

\paragraph{Encoder (Transformer backbone)}
The encoder consists of six Transformer blocks, each combining multi-head self-attention (MSA), a position-wise feed-forward network (FFN), residual connections, and layer normalization (LN), as illustrated by Figure 1. 
Formally, let $Z^{(l)} \in \mathbb{R}^{T \times d}$ denote the sequence of token embeddings entering the $l$-th layer:
\[
\hat{Z}^{(l)} = Z^{(l)} + \mathrm{MSA}\!\big(\mathrm{LN}(Z^{(l)})\big),
\]
\[
Z^{(l+1)} = \hat{Z}^{(l)} + \mathrm{FFN}\!\big(\mathrm{LN}(\hat{Z}^{(l)})\big).
\]

The attention sublayer aggregates contextual information across patches, while the FFN refines token representations independently. Stacking these blocks produces progressively more abstract and robust embeddings of the input sequence.

\paragraph{Decoder (MAE pipeline)}
The decoder mirrors the encoder structure. It restores the original sequence by inserting mask tokens, concatenating them with encoded visible embeddings, and predicting the missing patches.

\paragraph{Pretraining objective (MAE)}
Let $\hat{Y}$ denote decoder predictions and $Y$ the ground-truth patches. 
The objective is the mean squared error (MSE) over the masked positions $\mathcal{M}$:
\[
\mathcal{L}_{\mathrm{MAE}} = \frac{1}{|\mathcal{M}|}\sum_{i \in \mathcal{M}} \big\| \hat{Y}_i - Y_i \big\|_2^2.
\]

\paragraph{Downstream classification head}
For HAR fine-tuning, the MAE decoder is discarded and a lightweight classification head is attached to the encoder output. 
This head is a small multilayer perceptron (MLP) with BatchNorm and Dropout, projecting the encoder embedding dimension to $K$ activity classes. 
It is randomly initialized and trained during the fine-tuning phase.

\begin{figure}[!t]
    \centering
    \includegraphics[width=\columnwidth]{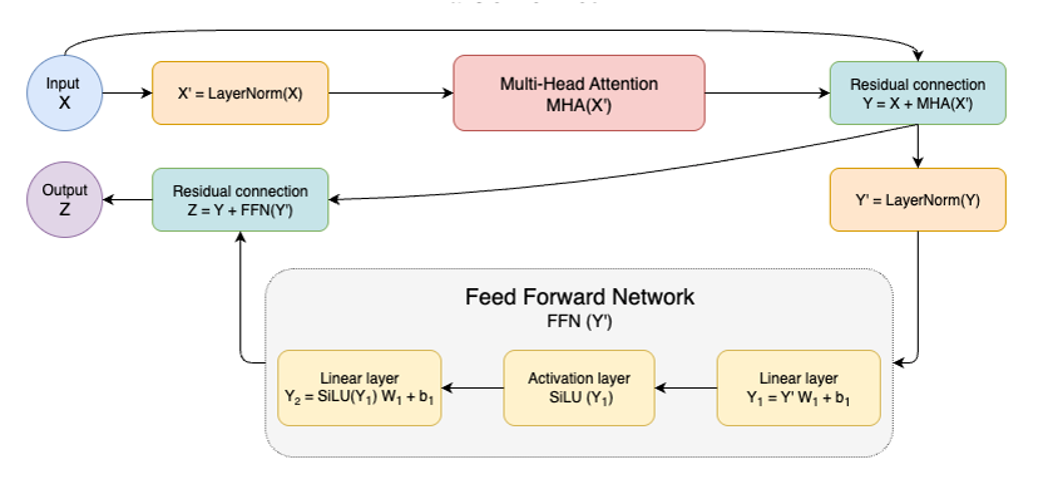}
    \caption{Detailed Encoder Architecture.}
    \label{fig:encoder_arc}
\end{figure}

\subsection{LoRA Integration}

LoRA adapters are inserted into the encoder’s linear projections, specifically within the feed-forward layers and the attention mechanism. 
The pretrained weight matrices remain frozen, while the adapters introduce a small number of trainable parameters. 
During fine-tuning, only these adapters and the classification head receive gradients, ensuring parameter efficiency without altering the backbone’s core architecture.

\subsubsection{LoRA in Feed-Forward Layers}

Within Transformer blocks, FFNs refine token representations through nonlinear transformations. 
A standard FFN is implemented as a two-layer structure alternating between linear projection and nonlinear activation:
\[
\text{FFN}(X) = \sigma(X W_1 + b_1) W_2 + b_2 ,
\]
where $W_1 \in \mathbb{R}^{d \times h}$ projects the embedding into a higher-dimensional hidden space $h$, $\sigma(\cdot)$ is a nonlinear activation, and $W_2 \in \mathbb{R}^{h \times d}$ maps back to the original dimension $d$.  

With LoRA, modifications are applied to the linear transformations. 
Instead of updating the full matrices $W_1$ and $W_2$, they are augmented with compact low-rank adapters:
\[
W_1' = W_1 + A_1 B_1, \quad 
W_2' = W_2 + A_2 B_2 ,
\]
where $A_i, B_i$ are trainable matrices with rank $r \ll \min(d,h)$.  

This design preserves the pretrained knowledge encoded in $W_1$ and $W_2$, while the adapters provide a lightweight mechanism for domain-specific adaptation. 
Since the nonlinear activations are left unchanged, the expressive properties of the original FFN are maintained, and the adapted output is seamlessly propagated to subsequent layers.  

\begin{table*}[t]
\centering
\caption{Cross-dataset recognition performance for different fine-tuning strategies. The dataset in the first column is the target used for fine-tuning; all other datasets are pooled to pre-train the model (LODO).}
\label{tab:recognition_results}
\setlength{\tabcolsep}{6pt}
\renewcommand{\arraystretch}{1.15}
\begin{tabular}{@{}l l c c c c@{}}
\toprule
\textbf{Dataset} & \textbf{Fine-tuning Method} & \textbf{Accuracy} & \textbf{F1 Macro} & \textbf{Precision} & \textbf{Recall} \\
\midrule
HHAR       & Full Fine-Tuning          & \textbf{0.981} & \textbf{0.980} & \textbf{0.981} & \textbf{0.981} \\
HHAR       & LoRA                      & 0.960 & 0.958 & 0.961 & 0.960 \\
HHAR       & QLoRA                     & 0.955 & 0.952 & 0.956 & 0.955 \\
\midrule
RealWorld  & Full Fine-Tuning          & \textbf{0.907} & \textbf{0.907} & \textbf{0.917} & \textbf{0.910} \\
RealWorld  & LoRA                      & 0.859 & 0.861 & 0.875 & 0.866 \\
RealWorld  & QLoRA                     & 0.860 & 0.861 & 0.875 & 0.866 \\
\midrule
PAMAP      & Full Fine-Tuning          & \textbf{0.853} & \textbf{0.854} & \textbf{0.867} & \textbf{0.863} \\
PAMAP      & LoRA                      & 0.829 & 0.825 & 0.855 & 0.821 \\
PAMAP      & QLoRA                     & 0.829 & 0.824 & 0.855 & 0.821 \\
\midrule
Sense      & Full Fine-Tuning          & \textbf{0.961} & \textbf{0.958} & \textbf{0.960} & \textbf{0.961} \\
Sense      & LoRA                      & 0.956 & 0.950 & 0.953 & 0.954 \\
Sense      & QLoRA                     & 0.955 & 0.949 & 0.952 & 0.953 \\
\midrule
UCI        & Full Fine-Tuning          & \textbf{0.968} & \textbf{0.969} & \textbf{0.970} & \textbf{0.971} \\
UCI        & LoRA                      & 0.952 & 0.953 & 0.955 & 0.956 \\
UCI        & QLoRA                     & 0.953 & 0.953 & 0.956 & 0.956 \\
\bottomrule
\end{tabular}
\end{table*}

\subsubsection{LoRA in Attention Mechanisms}

Self-attention layers project each token sequence $X \in \mathbb{R}^{n \times d}$ into queries, keys, and values:
\[
Q = X W_Q, \quad K = X W_K, \quad V = X W_V ,
\]
with projection matrices $W_Q \in \mathbb{R}^{d \times d_q}$, $W_K \in \mathbb{R}^{d \times d_k}$, and $W_V \in \mathbb{R}^{d \times d_v}$. 
The attention output is then:
\[
\text{Attention}(Q,K,V) = \text{softmax}\!\left(\frac{QK^\top}{\sqrt{d_k}}\right)V ,
\]
followed by a projection back to dimension $d$:
\[
\text{Output} = \text{Attention}(Q,K,V) W_O , \quad W_O \in \mathbb{R}^{d_v \times d}.
\]

With LoRA, each weight matrix $W$ is kept frozen and a low-rank update is added:
\[
W' = W + \Delta W, \quad \Delta W = A B ,
\]
where $A \in \mathbb{R}^{d \times r}$ and $B \in \mathbb{R}^{r \times k}$ are trainable matrices, reducing the number of trainable parameters from $d \times k$ to $r(d+k)$.  

Concretely, in the attention mechanism we obtain:
\[
[Q,K,V] = X\big([W_Q,W_K,W_V] + [A_QB_Q, A_KB_K, A_VB_V]\big),
\]
\[
\text{Output} = \text{Attention}(Q,K,V)\,(W_O + A_O B_O).
\]

Each pair $(A_\cdot, B_\cdot)$ thus represents a LoRA adapter inserted into the corresponding projection, enabling efficient adaptation while leaving the main Transformer backbone intact.

\subsection{Quantized LoRA Extension}

Quantized Low-Rank Adaptation (QLoRA) builds directly on the LoRA framework by combining low-rank adapters with quantization of the pretrained weight matrices. 
The motivation is to further reduce memory usage while retaining the representational power of large Transformer backbones, thereby enabling fine-tuning on hardware with resource constraints.  

In our implementation, all major projection layers of the encoder—namely the attention projections ($W_Q, W_K, W_V, W_O$) and the feed-forward projections ($W_1, W_2$)—are stored in a 4-bit quantized format. 
These quantized weights are frozen during adaptation, while the LoRA adapters ($A,B$) and the classification head are trained in higher precision (e.g., FP16 or BF16).  

Formally, let $W \in \mathbb{R}^{d \times k}$ denote a pretrained weight matrix. 
Instead of storing $W$ in full precision, it is compressed into a 4-bit representation $\tilde{W}$ via blockwise quantization. 
During forward propagation, $\tilde{W}$ is dequantized into an approximate floating-point representation $\hat{W}$, and the effective update is:
\[
W' = \hat{W} + \alpha \cdot (AB),
\]
where $A \in \mathbb{R}^{d \times r}$ and $B \in \mathbb{R}^{r \times k}$ are trainable low-rank matrices and $\alpha$ is a scaling factor. 
Gradients are backpropagated through $\hat{W}$, but optimizer updates are restricted to the LoRA adapters and the classification head. 
The quantized base weights $\tilde{W}$ remain fixed, preserving the knowledge encoded during large-scale pretraining.  

To maintain stability, certain components such as layer normalization parameters, positional embeddings, and patch embeddings are kept in higher precision. 
This precaution avoids numerical instabilities and ensures that quantization errors do not accumulate across layers. Overall, QLoRA provides a balance between three key requirements: preserving representational fidelity, reducing the memory footprint of frozen weights, and enabling deployability on memory-constrained devices such as wearables and IoT platforms.

\section{Results}

\subsection{Overview}

This section evaluates the empirical performance of the fine-tuning methods presented earlier. 
All experiments are conducted on the same MAE-Transformer backbone under a unified training setup and evaluation protocol to ensure comparability. 
Each model is fine-tuned for 50 epochs on five benchmark HAR datasets using a Leave-One-Dataset-Out (LODO) validation scheme. 
Within each target dataset, 70\% of the data is used for fine-tuning and 30\% is reserved for validation.

We analyze several practical aspects that are critical for real-world deployment:
\begin{itemize}
    \item Recognition quality using common metrics;
    \item Parameter efficiency: number of trainable parameters introduced by each method;
    \item Computational cost: training time measured in seconds;
    \item Memory usage: parameter storage and buffer memory during fine-tuning.
\end{itemize}



\subsection{Recognition Accuracy across Datasets}

In this section, we evaluate the recognition performance of each fine-tuning strategy across the five open-source datasets presented earlier (HHAR, RealWorld, PAMAP, Sense, and UCI). Each dataset presents a different level of difficulty due to variations in sensor modalities, sampling frequencies, user populations, and activity granularity. All models were evaluated in the LODO setting to simulate realistic domain shift scenarios, where the model must generalize to data collected from a previously unseen source. We report four commonly used classification metrics: Accuracy (overall proportion of correct predictions), Macro-F1 score (balance between precision and recall averaged across classes), Precision (proportion of correct positive predictions), and Recall (proportion of true positives correctly identified).

Table~\ref{tab:recognition_results} reports a detailed comparison of these metrics across datasets and fine-tuning methods. For each entry, the target dataset corresponds to the held-out domain in the LODO setting, i.e., the model is trained on the four remaining datasets and evaluated on the unseen one.
As expected, full fine-tuning consistently achieves the highest recognition accuracy and macro-F1 scores across all benchmarks. However, both LoRA and QLoRA perform competitively, often matching or closely approaching full fine-tuning performance, particularly on larger or less noisy datasets such as HHAR and UCI. Notably, QLoRA shows a slight underperformance compared to LoRA in certain settings (e.g., PAMAP), likely due to the approximation error introduced by quantization. Nevertheless, the performance gap remains relatively small (typically within 1–2\%), indicating that the temporal representations learned during MAE pretraining are sufficiently robust to tolerate low-bit adaptation mechanisms.

Overall, these results validate the effectiveness of parameter-efficient tuning strategies in HAR: despite significantly fewer trainable weights and reduced memory usage, both LoRA and QLoRA maintain strong classification performance across diverse domains.

\subsection{Trainable Parameters and Model Efficiency}

While recognition performance is essential for evaluating a fine-tuning method, it is equally important to consider its efficiency in terms of computational cost and memory usage—particularly in resource-constrained environments such as mobile or embedded HAR systems.
This subsection examines the parameter footprint and training overhead of each method to better highlight the trade-offs between adaptation quality and efficiency.
Specifically, we report:
\begin{itemize}
\item the number of trainable parameters and the total parameter count (trainable + frozen),
\item training time over 50 epochs (in seconds),
\item and peak memory consumption during fine-tuning (in megabytes).
\end{itemize}
Table~\ref{tab:params_summary} summarizes the number of trainable and total parameters required for each fine-tuning strategy. These values are architecture-dependent and remain constant across datasets, since all experiments rely on the same MAE backbone and identical low-rank adapter configurations (e.g., fixed rank $r$ for LoRA/QLoRA).

\begin{table}[H]
\centering
\begin{tabular}{|l|r|r|}
\hline
\textbf{Fine-tuning Method} & \textbf{Trainable Params} & \textbf{Total Params} \\
\hline
Full Fine-Tuning & 2,210,857 & 2,210,857 \\
LoRA             & 428,832   & 2,636,105 \\
QLoRA            & 428,832   & 2,636,105 \\
\hline
\end{tabular}
\caption{Trainable and total parameters for each fine-tuning strategy.}
\label{tab:params_summary}
\end{table}

Full fine-tuning updates all weights of the pretrained encoder and classification head, totaling approximately 2.2 million parameters. In contrast, LoRA and QLoRA freeze the pretrained weights and optimize only a small number of adapter parameters (about 428,000), yielding a fivefold reduction in the number of trainable weights.

Interestingly, the total parameter count for LoRA and QLoRA is slightly higher than for full fine-tuning. This occurs because LoRA introduces additional low-rank matrices ($A$, $B$) alongside the frozen pretrained weights $W$. Since these adapters are added as external components rather than replacing the original weights, the full-precision parameters are still retained.
In other words, LoRA and QLoRA extend the model architecture instead of overwriting existing weights. This explains why
$\text{Trainable}{\text{LoRA}} + \text{Total}{\text{Full}} > \text{Total}_{\text{LoRA}},$
as adapter weights function as parallel branches within selected layers (e.g., attention and feed-forward projections) rather than as direct substitutions.

\subsection{Memory consumption}

Next, we analyze memory consumption during adaptation. As shown in Table~\ref{tab:memory_and_params}, both LoRA and QLoRA allocate approximately 1.64~MB for trainable parameters, corresponding to the total size of the low-rank adapter matrices introduced by the fine-tuning strategy. The column Frozen Param (MB) quantifies the memory occupied by the pretrained encoder’s frozen weights, while Trainable Param (MB) indicates the size of the newly added trainable components. Together, these two quantities define the static memory footprint of the model parameters.

In practice, however, fine-tuning also incurs additional temporary costs, reported in the Buffer Memory (MB) column. This value captures the peak size of intermediate memory buffers required during training, including those for activations, gradients, and extra computations introduced by quantization mechanisms.

\begin{table}[H]
\centering
\begin{tabular}{|l|c|c|c|}
\hline
\textbf{Method} & \textbf{\begin{tabular}[c]{@{}c@{}}Frozen \\ Param (MB)\end{tabular}} 
                & \textbf{\begin{tabular}[c]{@{}c@{}}Trainable \\ Param (MB)\end{tabular}} 
                & \textbf{\begin{tabular}[c]{@{}c@{}}Buffer \\ Memory (MB)\end{tabular}} \\
\hline
LoRA    & 10.06 & 1.64 & 0.01 \\
QLoRA   & \textbf{6.22} & 1.64 & 4.82 \\
\hline
\end{tabular}
\caption{Memory consumption for LoRA and QLoRA.}
\label{tab:memory_and_params}
\end{table}

In the case of LoRA, buffer memory usage is negligible (0.01~MB), since all operations rely on standard floating-point arithmetic without additional processing of frozen weights.  
By contrast, QLoRA introduces a memory--computation trade-off: quantizing the frozen weights reduces their static footprint by about 40\% (from 10.06~MB to 6.22~MB), but requires on-the-fly dequantization during forward passes.  

In our CPU-based implementation, this process involves storing auxiliary scale factors and performing extra matrix operations to reconstruct floating-point tensors at runtime. As a result, buffer usage rises sharply to 4.82~MB---several hundred times higher than that of LoRA.  

This overhead largely reflects the limitations of our simplified implementation. Optimized GPU-based versions of QLoRA (e.g., with fused low-bit kernels and better memory management) handle dequantization far more efficiently, drastically reducing buffer costs and yielding substantially greater overall memory savings.

\subsection{Training time}

Training time is another key efficiency metric, as shown in Table~\ref{tab:training_time}.  
As expected, full fine-tuning consistently requires the longest time across datasets, owing to the need to update all model weights and maintain large optimizer states.  
LoRA generally achieves faster training than QLoRA, likely because QLoRA introduces additional overhead from quantization and dequantization during forward passes.  
Nevertheless, both methods provide substantial efficiency gains compared to full fine-tuning.

\begin{table}[H] \centering \begin{tabular}{|l|r|r|r|} \hline \textbf{Dataset} & \textbf{Full Fine-Tuning} & \textbf{LoRA} & \textbf{QLoRA} \\ \hline HHAR & 2772 & \textbf{2505} & 2598 \\ REALWORLD & 5657 & \textbf{5073} & 5195 \\ PAMAP & 1170 & \textbf{1011} & 1037 \\ SENSE & 1453 & \textbf{1299} & 1378 \\ UCI & 927 & 881 & \textbf{875} \\ \hline \end{tabular} \caption{Training time (in seconds) for each fine-tuning method across datasets.} \label{tab:training_time} \end{table}

It is important to note that total training time also depends on the size and complexity of the dataset.  
Datasets with more samples or longer input sequences require additional iterations per epoch and larger memory usage during training, which can substantially increase runtime.  
This explains the large variation in training time observed even within the same fine-tuning strategy across datasets. For instance, REALWORLD requires over 5600 seconds for full fine-tuning, whereas the smaller UCI dataset completes training in under 1000 seconds.  

Although LoRA reduces the number of trainable parameters by more than 80\%, the resulting decrease in training time is relatively modest (around 10\%).  
This limited speedup arises because LoRA does not reduce the number of operations during forward and backward passes: the frozen encoder still participates fully in computation, and partial backward graphs must be constructed for adapter training.  
In addition, LoRA introduces extra matrix multiplications, which partially offset its efficiency gains.  
In practice, the majority of training time is dominated by backbone computations and batch processing rather than parameter updates.  
Furthermore, training efficiency is influenced by several factors such as model depth, batch size, the number of normalization and activation layers, and the choice of optimizer—most of which remain unaffected by parameter sparsity.

\subsection{Impact of LoRA Rank on Accuracy and Efficiency}

To better understand the trade-offs between adaptation quality and efficiency in LoRA, we analyze the effect of the adapter rank---a key hyperparameter that determines the expressive capacity of the low-rank updates.  
The rank controls the dimensionality of the matrices $A$ and $B$ in each LoRA module and thus directly impacts both the number of trainable parameters and the computational cost of fine-tuning.  
Higher ranks increase the number of trainable weights and may enhance model expressiveness, whereas lower ranks reduce memory usage and accelerate training, but can limit performance.  

\begin{figure}[H]
\centering
\includegraphics[width=1\linewidth]{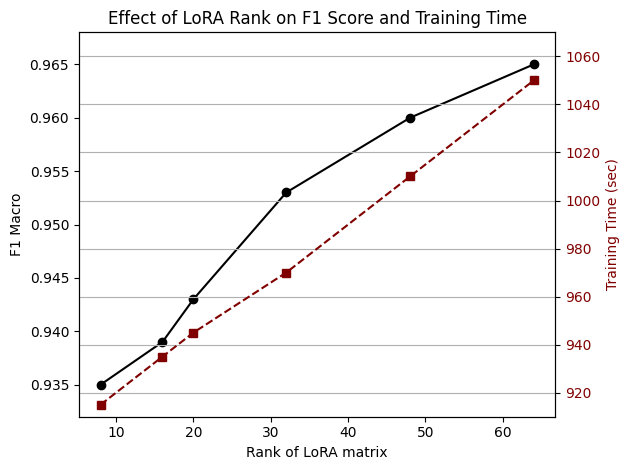}
\caption{F1 Macro and training time as a function of LoRA adapter rank.}
\label{fig:lora_rank_plot}
\end{figure}

To explore this trade-off, we conducted a focused experiment on the UCI dataset by fine-tuning the pretrained MAE encoder with LoRA adapters of varying ranks: $\{8, 16, 20, 32, 48, 64\}$.  
For each configuration, we measured the macro-averaged F1 score after 50 epochs of training, along with the total training time in seconds.  
The results are reported in Table~\ref{tab:lora_rank_effect} and illustrated in Figure~\ref{fig:lora_rank_plot}.

\begin{table}[H]
\centering
\begin{tabular}{|c|c|c|}
\hline
\textbf{LoRA Rank} & \textbf{F1 Macro} & \textbf{Training Time (sec)} \\
\hline
8  & 0.9352 & 916 \\
16 & 0.9395 & 934 \\
20 & 0.9434 & 945 \\
32 & 0.9531 & 971 \\
48 & 0.9602 & 1015 \\
64 & 0.9649 & 1049 \\
\hline
\end{tabular}
\caption{Effect of LoRA matrix rank on classification quality and training speed (UCI dataset).}
\label{tab:lora_rank_effect}
\end{table}

As expected, classification quality improves monotonically with increasing rank.  
Macro-F1 scores rise from 0.9352 at rank 8 to 0.9649 at rank 64, indicating that higher-rank adapters capture more nuanced task-specific representations.  
This gain, however, comes at the expense of training efficiency: total runtime increases from 916 seconds (rank 8) to over 1049 seconds (rank 64).  
These results highlight a clear trade-off: smaller ranks enable faster and more memory-efficient adaptation, whereas larger ranks improve accuracy at the cost of higher computation.  

Notably, performance gains saturate around ranks 48–64, suggesting diminishing returns beyond this point.  
In practice, moderate ranks (e.g., 32 or 48) may therefore provide the best balance between accuracy and efficiency, particularly in time- or memory-constrained deployment scenarios.  

Although this experiment was conducted only on the UCI dataset, the observed trend is likely to generalize to other HAR domains, given the shared temporal structure of sensor data.  
Overall, the findings emphasize the importance of tuning LoRA hyperparameters to match both task complexity and resource constraints.

\subsection{Robustness to Training Data Size}

Earlier in our evaluation, we showed that LoRA significantly reduces the number of trainable parameters compared to full fine-tuning.  
This opens the possibility of performing effective adaptation even when only a limited amount of labeled training data is available.

To explore this hypothesis, we fine-tuned models using various train/test splits, ranging from 70/30 to 30/70, and recorded the resulting accuracy on the test set.  
The results are summarized in Table~\ref{tab:accuracy_split}.

\begin{table}[H]
\centering
\begin{tabular}{|c|c|c|c|}
\hline
\textbf{Split} & \textbf{Full FT Accuracy} & \textbf{LoRA Accuracy} & \textbf{LoRA / Full FT} \\
\hline
70/30 & 0.9676 & 0.9528 & 0.9847 \\
60/40 & 0.9651 & 0.9508 & 0.9852 \\
50/50 & 0.9618 & 0.9483 & 0.9859 \\
40/60 & 0.9543 & 0.9416 & 0.9867 \\
30/70 & 0.9380 & 0.9257 & 0.9869 \\
\hline
\end{tabular}
\caption{Accuracy comparison under different train/test splits.}
\label{tab:accuracy_split}
\end{table}

As expected, both full fine-tuning and LoRA exhibit a gradual decline in recognition accuracy as the training set becomes smaller. However, the drop in LoRA accuracy is relatively modest compared to full fine-tuning, and the ratio of LoRA to full fine-tuning performance (Table~\ref{tab:accuracy_split}) remains remarkably stable, even improving as the training size decreases.

This suggests that LoRA maintains a higher degree of data efficiency: it generalizes well even with fewer training examples.  
In practical scenarios where collecting large-scale labeled data is expensive or impractical, LoRA offers a compelling alternative that delivers strong performance with fewer parameters and less supervision.

\subsection{Synthesis}

In summary, both LoRA and QLoRA demonstrate substantial efficiency gains over full fine-tuning, while preserving competitive recognition performance.  
As expected, LoRA significantly reduces the number of trainable parameters and achieves faster training times by freezing the pretrained backbone and optimizing only lightweight adapter layers.  
This makes it particularly attractive for rapid model adaptation in resource-constrained environments, as well as for few-shot learning scenarios where only limited labeled data is available.  

QLoRA builds upon this foundation by quantizing the frozen weights, thereby further reducing the static memory footprint without degrading classification quality.  
Although our implementation shows a modest increase in buffer memory, QLoRA maintains the same number of trainable parameters as LoRA while substantially lowering the size of frozen weights.  
This makes QLoRA especially well suited for deployments where memory capacity is the primary bottleneck, such as on low-power IoT devices.

\section{Conclusion}

In this paper, we addressed the challenge of adapting deep learning models for human activity recognition (HAR) under limited computational and memory resources. Since full fine-tuning is often impractical in real-world deployments, we investigated parameter-efficient alternatives—LoRA and QLoRA—against the full fine-tuning baseline. All methods were evaluated on the same transformer-based backbone using a Leave-One-Dataset-Out protocol across five HAR benchmarks, simulating deployment to unseen domains. Results show that LoRA and QLoRA achieve competitive performance, typically within 1–2\% of full fine-tuning, while requiring far fewer trainable parameters. Importantly, LoRA also reduces the amount of labeled data needed to adapt a pretrained model, enabling effective fine-tuning under scarce supervision. LoRA reduced trainable weights more than fivefold and improved training time, while QLoRA further lowered memory usage by about 40\% through quantization of frozen weights.

Additional analyses revealed that LoRA’s accuracy improves with adapter rank up to a plateau around rank 32, and that it remains robust under reduced supervision, retaining over 98\% of full fine-tuning accuracy even with limited labeled data; in practice, this lowers the labeled-data requirement to reach a target accuracy. These findings confirm that LoRA and QLoRA offer scalable and data-efficient personalization strategies for HAR.

Future directions include extending evaluation to CNN or hybrid backbones, exploring online and continual learning settings, and automating adapter/quantization configurations based on deployment constraints. Another ongoing line of work is to integrate LoRA into a federated learning pipeline, used in a hybrid fashion with conventional fine-tuning (e.g., on selected layers or rounds) to limit client divergence (client drift) and to enable faster global model convergence \cite{EK2022101714}


\bibliographystyle{IEEEtran}
\bibliography{sample-base}


\end{document}